\documentclass{article}
\usepackage{spconf,amsmath,graphicx}
\usepackage[american]{babel}
\usepackage{microtype}
\usepackage{url}

\newcommand{\etal}{\textit{et~al.\,}}

\newcommand{\ie}{\textit{i.e.}}

\title{Curriculum Knowledge Switching for Pancreas Segmentation}
\name{Yumou Tang, Kun Zhan$^\star$, Zhibo Tian, Mingxuan Zhang, Saisai Wang, Xueming Wen}
\address{School of Information Science and Engineering, Lanzhou University\\
\url{https://github.com/kunzhan/CKS_Pancreas}}
\begin{document}
%
\maketitle
\begin{abstract}
Pancreas segmentation is challenging due to the small proportion and highly changeable anatomical structure. It motivates us to propose a novel segmentation framework, namely Curriculum Knowledge Switching (CKS) framework, which decomposes detecting pancreas into three phases with different difficulty extent: straightforward, difficult, and challenging. The framework switches from straightforward to challenging phases and thereby gradually learns to detect pancreas. In addition, we adopt the momentum update parameter updating mechanism during switching, ensuring the loss converges gradually when the input dataset changes. Experimental results show that different neural network backbones with the CKS framework achieved state-of-the-art performance on the NIH dataset as measured by the DSC metric.
\end{abstract}
\section{Introduction}\label{sec:intro}

This paper focuses on pancreas segmentation from CT-scanned images, which is a crucial step in diagnosing and supporting pancreatic cancer surgery. The segmentation of the pancreas is challenging for two basic reasons: (1) The pancreas only takes up about 1.5\% of a typical 2D abdominal multi-organ CT image, which means that both size and resolution have posed significant challenges to the segmentation network; (2) In addition, unlike the heart, kidneys, and other organs, whose geometric shapes are essentially fixed, the pancreatic organs of different people differ greatly.
\begin{figure}[tb]
  \centering
  \label{motivation}
  \includegraphics[width=0.45\textwidth]{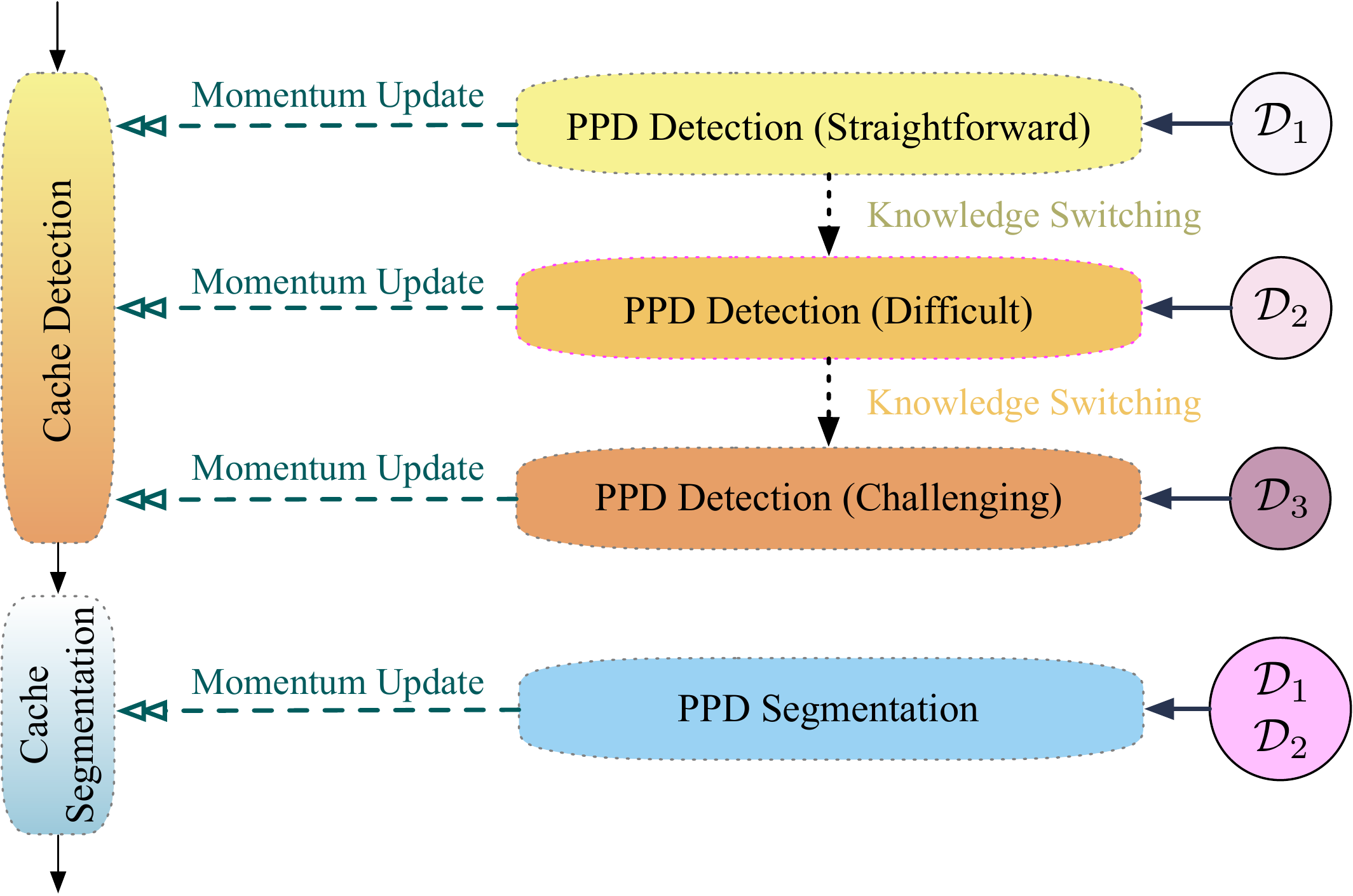}\\
  \caption{Overview of our proposed framework. Several Pancreatic Pixel Discriminators (PPD) are built to progressively switch from straightforward, difficult, and challenging datasets, $\mathcal D_1$, $\mathcal D_2$, and $\mathcal D_3$. Learned knowledge is stored in the two cache models by the momentum update mechanism and is switched to the subsequent PPDs.}
\end{figure}

Most contemporary approaches employ the coarse-to-fine framework~\cite{Roth2015DeepOrgan, farag2016bottom, Zhou2017A, yu2018recurrent}, in which the coarse network uses the original image as input to determine the approximate pancreatic position and produces a bounding box. The created bounding box is then used to locate the pancreas and crop the original image to obtain a smaller cropped image with a larger pancreas-to-background ratio, followed by the fine network, which is used to segment the cropped image with good precision. The two-stage technique trains two distinct networks separately on large and small images, resolving the issue that the pancreatic content of the original image is too small to segment directly. However, the method is effective but heavily dependent on the detection results of the coarse model. Recent efforts tend to directly apply the raw dataset to train the coarse network, and we argue that it is challenging for the model to learn representation on the raw dataset properly because the number of pancreatic pixels in a CT picture is too few to be resolved by the coarse network alone.

According to this observation, we propose a novel Curriculum Knowledge Switching framework inspired by curriculum learning \cite{bengio2009curriculum}. Instead of directly mapping the input domain to the label domain, we employ four specialized training phases with distinctive datasets $\mathcal D_1, \mathcal D_2$, and $\mathcal D_3$ and neural networks to decompose the learning of pancreatic patterns into smaller components. The framework progressively switches from $\mathcal D_1$ to $\mathcal D_3$ in order to learn from detecting the pancreatic region properties to extracting the pancreatic-specific pixels. In order to effectively inherit knowledge learned in previous stages, we employ the momentum update mechanism to incrementally switch the learned knowledge to two cache models, as shown in Fig.~\ref{motivation}.

Experimental results on the NIH dataset reveal that our framework with multiple backbone types outperforms state-of-the-art approaches based on the DSC metric. The CKS framework incrementally enhances segmentation performance and identifies the global optimum of precise pancreas segmentation.

The main contributions of this paper are listed as follows:
\begin{itemize}
\item We propose a Curriculum Knowledge Switching (CKS) framework to teach neural networks constructing pancreas discriminate ability in a consecutive manner, which starts from straightforward to challenging problems.
\item The momentum update method is used for knowledge switching, ensuring the learned knowledge from previous stages inherits to the following tasks.
\item We employ a data augmentation strategy to refine the learning of the segmentation stage. Experimental results have shown the superiority of the proposed frameworks across different backbones.
\end{itemize}
\section{Methodology}\label{sec:format}
\subsection{The Curriculum Knowledge Switching Framework}
We observe that semantic segmentation models for the pancreas could be regarded as Pancreatic Pixel Discriminators (PPDs), which are utilized to estimate pancreatic pixel locations while filtering out background pixels. Due to the varying shape of the pancreas organ, accurately detecting the pancreas poses great challenges. We construct units with suitable levels of complexity to overcome this issue, enabling the PPDs to gradually learn the location and specific pixels of the pancreas.

In this work, we set three phases with varying difficulty levels, straightforward, difficult, and challenging, to progressively guide the framework to identify the pancreas from the complicated CT scan background, as shown in Fig.~\ref{kt}. These three steps correspond to three training sets: the ground-truth bounding box cropped from the label set ($\mathcal D_1$) dominated by pancreatic pixels; the cropped bounding box predicted by the coarse model ($\mathcal D_2$), which has more background than $\mathcal D_1$ and contains boundary ambiguity pixels; and the raw image ($\mathcal D_3$), which is the raw CT slices and is mainly occupied by the background pixels. At each level, we assign a distinctive PPD to map the corresponding domains.

Since the model of the detection stage learns the requisite pancreatic characteristics, we wish to switch the learned knowledge from the detection stage to the segmentation stage during the succeeding segmentation step.
\subsubsection{Momentum Update for Knowledge Transferring}
As we utilize multiple PPDs to locate and learn precise features of the pancreas, retaining the performance of the trained model when switching between phases is necessary. We implement two methods for effective knowledge switching. One is to directly copy the trained model to the second one, and the other is to incrementally collect the acquired information, introduced as follows:

A deep neural network model is denoted by $f(X|\Theta)$, where $X$ is the input image and $\Theta$ is the model weight parameters, updated by the back-propagation algorithm. The momentum update approach gradually updates the model parameters $\Theta_{\rm mu}$ from $\Theta$ by:
\begin{equation}
\Theta_{\rm mu}=m\Theta_{\rm mu}+(1-m)\Theta
\end{equation}
where $m\in[0,1]$ is a momentum coefficient that defines to what extent the historical models should influence the current model. Accordingly, the momentum update method is able to update $\Theta_{\rm mu}$ smoothly \cite{he2020momentum}.


\subsubsection{Detection Stage}
\begin{figure}[tb]
\begin{minipage}[b]{1\linewidth}
  \centering
  \centerline{\includegraphics[width=8.6cm]{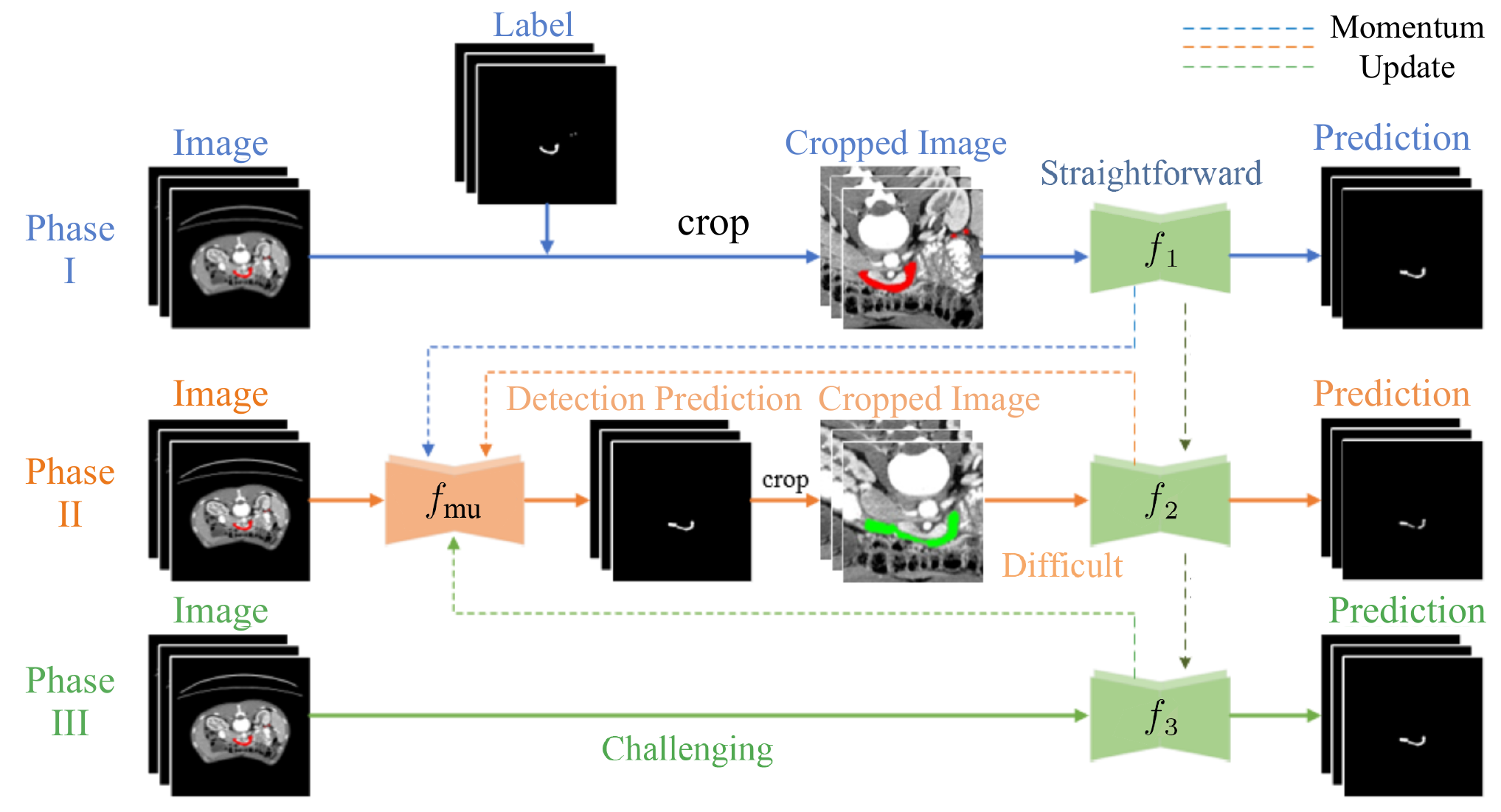}}
  \caption{The detection stage of the proposed framework. First, $f_1$ is trained by the image cropped by its label in $\mathcal D_1$. Second, the detection cache model $f_{\rm mu}$ is utilized to produce the training set $\mathcal D_2$ and the $f_2$ is further trained by datasets in $\mathcal D_2$. Lastly, the PPD $f_3$ is trained by the image in $\mathcal D_3$.}\medskip
  \label{kt}
\end{minipage}
\end{figure}

From Fig.~\ref{kt}, the detection stage of our CKS system employs three PPDs $\{{f_1, f_2, f_3}\}$ and three training sets $\{{\mathcal D_1, \mathcal D_2, \mathcal D_3}\}$. The PPDs switch from $\mathcal D_1$ to $\mathcal D_3$ and acquire the ability to dectect the pancreas. In addition, we build an extra cache model, namely Detection Cache, to accumulate the knowledge learned in the three phases with the momentum update mechanism, making it more robust and stable than all three PPDs.

\textbf{Phase I} \quad In $\mathcal D_1$, the proportion of the pancreas is relatively large, and the positive and negative examples are well balanced. We employ a PPD here, namely $f_1$ to learn detailed patterns of the pancreas. In this phase, the input $X\in \mathcal D_1$ is cropped by utilizing the ground-truth $T$ to generate image $X'$ and we feed $X'$ to $f_1$,
\begin{equation}
\begin{cases}
X'={\rm crop}(X, T)\in \mathcal D_1\\
P_{1} = f_1(X'|\Theta_1)
\label{eq1}
\end{cases}
\end{equation}
where the first PPD is denoted by $f_{1}(X'|\Theta_{1})$, $P_1$ is the prediction of $f_1$, and $\Theta_{1}$ is its parameters, updated by the back-propagation algorithm. The parameters of the detection cache, $\Theta_{\rm mu}$, are updated by the momentum update:
\begin{equation}
\Theta_{\rm mu}=\alpha \Theta_{\rm mu} + (1-\alpha)\Theta_{1}\,\label{mucm}
\end{equation}
After the first phase, $f_{\rm mu}(X|\Theta_{\rm mu})$ is able to identify the overall shape of the pancreas and yet failed to extract pancreas features from datasets with low pancreas pixels ratio. This issue is further addressed in the second phase.

\textbf{Phase II} \quad  In this phase, the detection cache model produces candidate frames with more boundary ambiguity pixels and a larger percentage of non-pancreatic pixels. The produced training set is $\mathcal D_2$. Using the pancreatic pixel discrimination capabilities of $f_{\rm mu}$, the raw picture $X\in \mathcal D_3$ is cropped to generate $X''$.
\begin{equation}
\begin{cases}
P_{\rm mu} =f_{\rm mu}(X|\Theta_{\rm mu}), \\
X''={\rm crop}(X,P_{\rm mu}>0.5)\in \mathcal D_2\\
P_2=f_2(X''|\Theta_2)\\
\Theta_{\rm mu}=\alpha \Theta_{\rm mu} + (1-\alpha)\Theta_{2}	
\end{cases}
\end{equation}

With $\mathcal D_2$, we employ a PPD, $f_2$, with weights initialized by $\Theta_{1}$ to guide the framework switching to the difficult phase of learning. We argue that switching from $\mathcal D_1$ to $\mathcal D_2$ bridges a critical step to guarantee the final detection accuracy.

\textbf{Phase III} \quad In the third phase, \ie, the challenging phase, we intend to guide the framework to extract pancreatic features directly from the raw CT scan input slices  $\mathcal D_3$. We employ a PPD, $f_3$, with weights initialized by $\Theta_{2}$. This step renders the $f_3$ consistent with the real situation and obtains high pancreatic pixel discriminative ability. The learned knowledge is switched and stored in the Detection Cache model with the momentum update mechanism. This phase is formulated as:
\begin{equation}
	\begin{cases}
P_{3} =f_{_3}(X|\Theta_{3}), \\
\Theta_{\rm mu}=\alpha \Theta_{\rm mu} + (1-\alpha)\Theta_{3}
	\end{cases}
\end{equation}

Notably, our strategy is heavily based on the momentum update mechanism, which guarantees effective knowledge switching. The detailed pancreatic properties learned during the first phase of training are switched to $f_2$ in the second phase, passing the first phase categorization capabilities to the second, and the momentum update mechanism provides additional information regarding the peripheral characteristics of the pancreas to $f_2$ during the second phase. In the third phase, dataset $\mathcal D_3$ enhances the model PPD ability to generalize from real-world scenarios. The entire training process is dynamic, establishing not only a balance between the detection and segmentation stages but also high performance.
\subsubsection{Segmentation Stage}
This stage aims to precisely segment the pancreas area according to the detection of the previous stage. The pixel discrimination abilities vary at various stages of the detection training. However, following the completion of the three-step coarse training, the Detection Cache model $f_{\rm mu}$ is fixed, and its prediction behaves identically to the actual test circumstance. Therefore, we combine $\mathcal D_2$ into $\mathcal D_1$ to train the segmentation model $f^{\rm seg}$, allowing the framework to switch from detecting the pancreas to extracting the detailed features of the pancreas. The segmentation model parameters are also stored in the Segmentation Cache Model $f^{\rm seg}_{\rm mu}$with the momentum update, which is typically more reliable and smoother than $f^{\rm seg}$\,.
\subsection{Prediction}
The training stage yields two networks: the Cache Detection Model $f_{\rm mu}$ and the Cache Segmentation Model $f^{\rm seg}_{\rm mu}$. The prediction workflow is as follows: First, we feed the test image $X$ forward: $P_{\rm det}=f_{\rm mu}(X|\Theta_{\rm mu})\,.$
Second, we crop image $X$ to produce $X''\in \mathcal D_2$ by $X$, $X''={\rm crop}(X,P_{\rm det}>0.5)\,.$
Third, we feed $X''$ forward: $P_{\rm seg}=f^{\rm seg}_{\rm mu}(X''|\Theta^{\rm seg}_{\rm mu})\,.$
Fourth, we do thresholding on $P_{\rm seg}$ to obtain the segmentation result\,.
\subsection{Supervision}

In this paper, we adopt the intersection-over-union loss for model learning supervision:
\begin{equation}
\mathcal L_{\rm iou}=1-\frac{\sum_{ij}t_{ij}p_{ij}}{\sum_{ij}t_{ij}+\sum_{ij}p_{ij}-\sum_{ij}t_{ij}p_{ij}}\,.
\end{equation}
where $T=[t_{ij}]$ denotes the ground-truth binary label matrix and $P=[p_{ij}]=f(X|\Theta)$ is the prediction probability of pancreatic pixels.

Since the Pancreatic Pixel Discriminator is regarded as the binary classifier, we add the binary cross-entropy loss for better supervision:
\begin{equation}
\mathcal L_{\rm bce}=-\sum_{ij}\bigl(t_{ij}\ln p_{ij}+(1-t_{ij})\ln(1-p_{ij})\bigr)\,.
\end{equation}

As the adjacent pancreatic pixels have a high probability of being pancreatic area~\cite{muller2019does}, we use Gaussian kernel smoothed $\hat T=[\hat t_{ij}]$ and $\hat P=[\hat p_{ij}]$ to define a new loss, and it is formulated as:
\begin{equation}
\mathcal L_{\rm s}=1-\frac{1}{N}\sum_{ij}\frac{2\hat t_{ij}\hat p_{ij}}{\hat t_{ij}^2+\hat p_{ij}^2}
\end{equation}
where $N$ is the number of pixels of an image.

The overall loss is given by:
\begin{equation}
\mathcal L(P,T) = \mathcal L_{\rm iou} +\mathcal L_{\rm bce} +\mathcal L_{\rm s}\,.\label{loss}
\end{equation}

\section{Experimental Results}\label{sec:pagestyle}
\subsection{Experimental Setting}
We validate the CKS framework using the NIH pancreas dataset~\cite{Roth2015DeepOrgan} consisting of 82 contrast-enhanced abdominal CT volumes. Each CT volume is cut into a set of 2D slices and each slice is cut along three axes, \ie, the coronal, sagittal, and axial views. Following \cite{Zhou2017A}, experiments are conducted in a cross-validation manner on four-folds. The test set contains 20 images and each image has at least 5,000 valid slices in both directions, and thereby the reliability of our results is guaranteed. Finally, the Dice-S\o rensen coefficient (DSC) is exploited to measure performance.

\subsection{Detection Stage Performance}
\begin{figure}[tb]
\centering\includegraphics[width=0.47\textwidth]{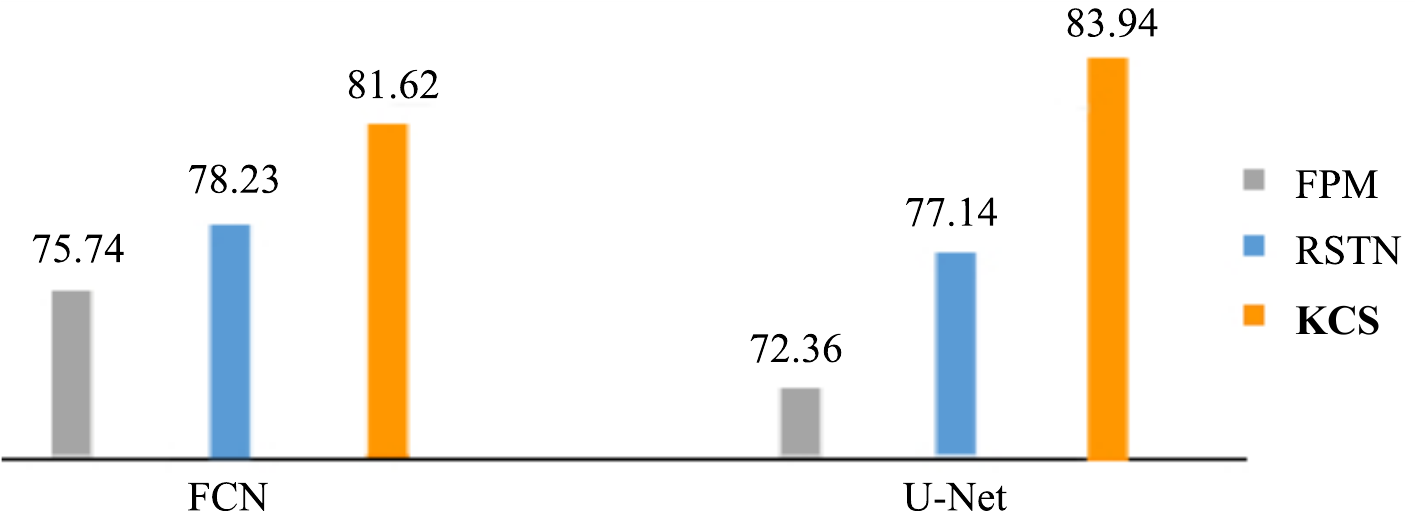}\\
\caption{Results of the detection stage of our proposed \textbf{CKS} framework comparing to FPM~\cite{Zhou2017A} and RSTN~\cite{yu2018recurrent} with two backbones (FCN-8s~\cite{Shelhamer2017Fully} and U-Net~\cite{ronneberger2015u}).}\label{CM_Res}
\end{figure}

According to Fig.~\ref{CM_Res}, the detection stage results of CKS outperform the performance of FPM~\cite{Zhou2017A} and RSTN~\cite{yu2018recurrent} under the same baseline, which is beneficial to the subsequent segmentation stage.
\begin{figure}[tb]
\centering\includegraphics[width=0.48\textwidth]{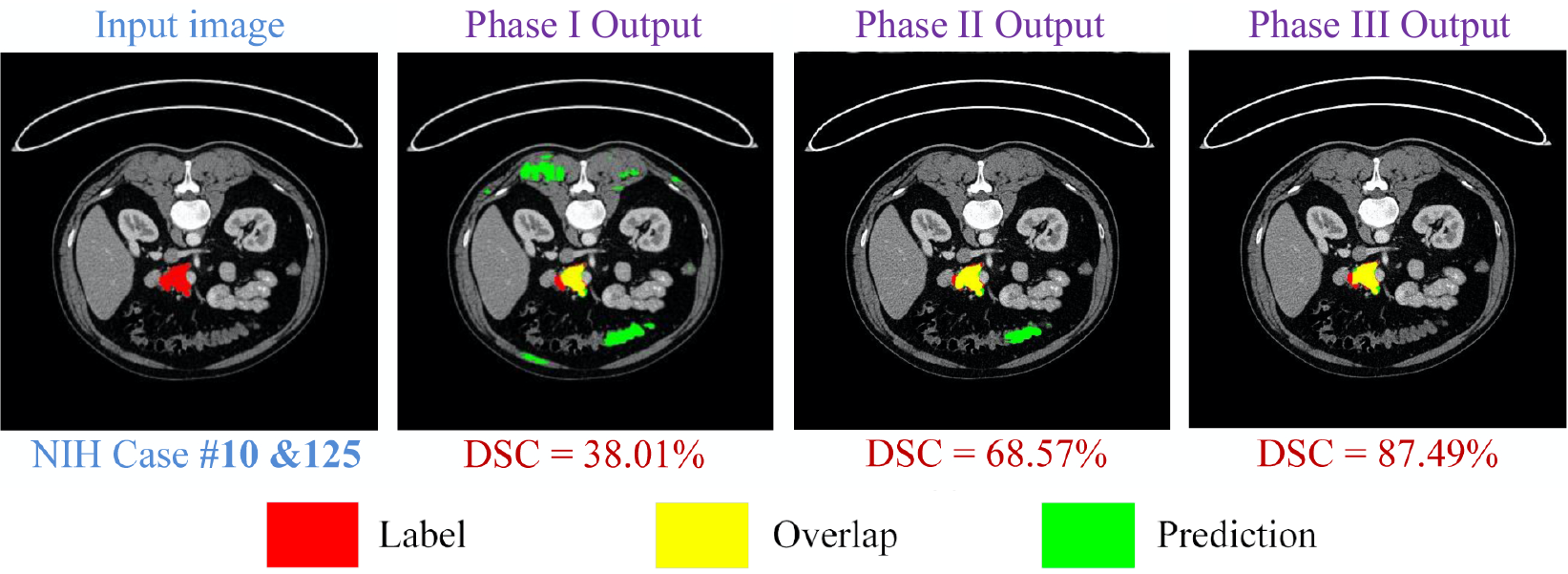}\\
\caption{The detection stage of \textbf{CKS} adapts real domain $\mathcal D_3$ gradually. Note that the prediction error (highlighted in green color) decreases as the framework switches to the challenging phase, along with the improvement of the DSC metric.}\label{CM_res}
\end{figure}

Fig.~\ref{CM_res} depicts the process of the proposed CKS framework progressively switching to the final detection result across three difficulty levels in detail. As the figure shows, the CKS framework guides the PPDs switching from $\mathcal D_1$ to $\mathcal D_3$ with the DSC metrics increasing steadily. Experiments reveal that training set $\mathcal D_2$ contains more non-pancreatic pixels than $\mathcal D_1$ and numerous pixels with border ambiguity data. The second training phase is closer to the real-world situation than $\mathcal D_2$ and gives more attention to ambiguous locations.

The detection result without the constructed first and the second phases show inferior performance, which are 81.33\% and 81.59\% in terms of the DSC metric, respectively, as compared to the result of 85.42\% with three full training phases.

\subsection{Overall Performance}
This section reports the prediction results. We utilize U-net as our network backbone in line with methods listed in Table~\ref{com_tlb}.

Following \cite{Zhou2017A}, DSC metrics of the detection model and the overall model under different thresholds are summarized in Table \ref{CTS_tlb}. The comparison results are provided in Table~\ref{com_tlb}. For the sake of a fair comparison, the metrics are from their papers or reproducing the publicly available code. The outcome demonstrates improvement over the state-of-the-art algorithms.
\begin{table}[tb]
\centering
\caption{DSC performance of the proposed \textbf{CKS} algorithm with thresholds(\%) following \cite{Zhou2017A}.}\label{CTS_tlb}
\begin{tabular*}{0.48\textwidth}{@{\extracolsep{\fill}\,}l|ccc}
\hline
Method & Mean & Max & Min \\
\hline
Detection & 83.94 & 91.01 & 58.45 \\
\hline
$d_t > 0.95$ & $\textbf{85.42}\pm\textbf{4.39}$ & 91.57 & 67.59\\
\hline
$d_t > 0.98$ & $85.41\pm4.44$ & 91.58 & 67.38 \\
\hline
$d_t > 0.99$ & $85.37\pm4.43$ & 91.61 & 66.50 \\
\hline
\end{tabular*}
\end{table}
\begin{table}[tb]
\centering
\caption{DSC performance of \textbf{CKS} and SOTA methods (\%). Methods with FCN-8s are above the double lines and the ones with U-Net are below the double lines.}\label{com_tlb}
\begin{tabular*}{0.48\textwidth}{@{\extracolsep{\fill}\,}l|ccc}
\hline
Method & Mean & Max & Min \\
\hline
DeepOrgan~\cite{Roth2015DeepOrgan} & 71.42$\pm$10.11 & 86.29 & 23.99\\
\hline
Roth \etal~\cite{2016Spatial} & 78.01$\pm$8.20&88.65&34.44\\
\hline
FPM~\cite{Zhou2017A} & 82.37$\pm$5.68 &90.85&62.43\\
\hline
Roth \etal~\cite{2018Spatial} & 81.27$\pm$6.27&88.96&50.69\\
\hline
Cai~\etal~\cite{cai2017improving}&82.40$\pm$6.70&90.10&60.00\\
\hline
RSTN~\cite{yu2018recurrent} & 84.50$\pm$4.97&91.02&62.81\\
\hline
Zhu~\etal~\cite{zhu20183d} & $84.59 \pm 4.86$ & 91.45& 69.92\\
\hline
Zhang~\etal~\cite{zhang2021deep} & $84.47 \pm 4.36$ & 91.54& \textbf{70.61}\\
\hline
STFFM~\cite{zhang2021automatic} & $84.90$ & 91.46& 61.82\\
\hline
\hline
U-Net~\cite{ronneberger2015u}  &$79.70 \pm 7.60$ & 89.30& 43.40\\
\hline
Attention U-Net~\cite{oktay2018attention}  & $83.10 \pm 3.80$ & 89.38& 66.77\\
\hline
\textbf{CKS}& $\textbf{85.42}\pm4.39$ & \textbf{91.57} & 67.59\\
\hline
\end{tabular*}
\end{table}


\section{Conclusion}\label{sec:typestyle}
We present a novel Curriculum Knowledge Switching (CKS) framework that directs the Pancreas Pixel Discriminators with designed adaptive phases to incrementally switch from easy to difficult training sets and acquire the capacity to localize and segment the pancreas gradually. The augmented dataset $\mathcal D_2$ is employed to train the second phase of the detection stage, bridging the direct switching from $\mathcal D_1$ to $\mathcal D_3$. Moreover, we utilize the momentum update to facilitate progressive knowledge switching, inheriting the learned knowledge to the subsequent ones. Experimental results demonstrate that CKS surpasses the state-of-the-art accuracy and is helpful for computer-assisted clinical diagnosis and detection of small objects.
\bibliographystyle{IEEEbib}
\bibliography{strings}
\end{document}